\documentclass[a4paper, 11pt]{article}
\usepackage[utf8]{inputenc}

\usepackage{import}
\usepackage{dsfont}
\usepackage{amsmath}
\usepackage{amssymb}
\usepackage{amsfonts}
\usepackage{tabularx}
\usepackage{mathtools}
\usepackage[left=2.5cm,right=2.5cm,top=3.5cm,bottom=3.5cm]{geometry}

\usepackage{algorithm}
\usepackage{algpseudocode}

\usepackage{color,colortbl}

\definecolor{cristianoorange}{rgb}{0.9,0.3,0.}
\definecolor{paolopink}{rgb}{0.9,0.3,0.9}

\usepackage[T1]{fontenc}    
\usepackage[unicode]{hyperref}       
\usepackage{url}            
\usepackage{booktabs}       
\usepackage{amsfonts}       
\usepackage{nicefrac}       
\usepackage{microtype}      
\usepackage{xcolor}         

\usepackage{authblk}
\usepackage[normalem]{ulem}

\title{Learning fast while changing slow in spiking neural networks}

\author[1]{Cristiano Capone}
\author[2]{Paolo Muratore}
\affil[1]{Natl. Center for Radiation Protection and Computational Physics, Istituto Superiore di Sanità, Rome, Italy}
\affil[2]{SISSA, Trieste, Italy}

\date{}

\begin{document}

\maketitle

\begin{abstract}
Reinforcement learning (RL) faces substantial challenges when applied to real-life problems, primarily stemming from the scarcity of available data due to limited interactions with the environment. This limitation is exacerbated by the fact that RL often demands a considerable volume of data for effective learning. The complexity escalates further when implementing RL in recurrent spiking networks, where inherent noise introduced by spikes adds a layer of difficulty.
Life-long learning machines must inherently resolve the plasticity-stability paradox. Striking a balance between acquiring new knowledge and maintaining stability is crucial for artificial agents.
To address this challenge, we draw inspiration from machine learning technology and introduce a biologically plausible implementation of proximal policy optimization, referred to as lf-cs (learning fast changing slow). Our approach results in two notable advancements: firstly, the capacity to assimilate new information into a new policy without requiring alterations to the current policy; and secondly, the capability to replay experiences without experiencing policy divergence.
Furthermore, when contrasted with other experience replay (ER) techniques, our method demonstrates the added advantage of being computationally efficient in an online setting.
We demonstrate that the proposed methodology enhances the efficiency of learning, showcasing its potential impact on neuromorphic and real-world applications.

\end{abstract}

\section*{Introduction}

The last decade has witnessed many advances in reinforcement learning (RL), with large efforts to provide biologically inspired implementations.
Authors in \cite{mnih2015human}, merging together reinforcement learning and deep learning, achieved competitive results in Atari games from pixel for the first time beginning the revolution of reinforcement learning for real-life problems. 

Deep reinforcement learning (DRL) algorithms such as Deep Q-Network (DQN) and Twin-Delayed Deep Deterministic Policy Gradient (TD3) for discrete and continuous action space environments have also been implemented in spiking networks \cite{patel2019improved,tang2021deep,akl2023toward}, showcasing their potential for addressing complex continuous control problems with state-of-the-art DRL techniques and highlighting avenues for reducing sensitivity to spiking neuron hyperparameters, ultimately enabling direct execution on neuromorphic processors for increased energy efficiency.
 
The advantages of spike-based systems are well-documented, including spike-driven weight updates and spike-driven communication between neurons. While this is advantageous for optimizing simulations in-silico, its significance is magnified exponentially in the context of neuromorphic hardware.

However, the training algorithms described above relies on non-local learning rules, e.g. on the combination of back-propagation (which is not local) and surrogate gradient \cite{zenke2021remarkable}, making the process computationally expensive and biologically implausible.
 
The major requirement for biological plausibility lies in the locality in both time and space. This necessity is fundamental, as without it, implementation in a physical system/hardware like neuromorphic chips, would be unattainable. Furthermore, locality guarantees energetic efficiency.

Also, methods that address biologically plausible learning rule (local in space and time) in biologically plausible architectures (recurrent spiking networks) \cite{florian2007reinforcement,fremaux2013reinforcement,bellec2020} have been proposed.
In particular \cite{florian2007reinforcement,fremaux2013reinforcement} introduced methods utilizing reward-based local plasticity rules, highlighting their effectiveness for simple tasks but limitations in complex robotic control tasks due to optimization constraints \cite{fremaux2013reinforcement}.
Authors in \cite{bellec2020} proposed e-prop, a biologically plausible implementation of actor-critic and backpropagation through time, which is deemed to be the state of the art in RL in spiking networks, and succeeding in relevant benchmarks such as atari games. Interestingly recent works on spiking architectures \cite{bellec2020,stockl2021optimized} demonstrates the possibility to achieve performances that are comparable to non spiking ones.

However, the challenge of effectively utilizing limited online data remains a fundamental concern. This raises the difficulty of managing online learning processes \cite{jimenez2014stochastic,gilra2017predicting,bellec2020}, especially when constrained by the availability and quality of data.
In the context of implementing reinforcement learning within recurrent spiking networks this becomes particularly challenging, as the intrinsic noise generated by spikes adds a further level of complexity, necessitating specialized strategies to stabilize output behaviors.
\cite{zenke2021remarkable,capone2022error,muratore2021target}.
Furthermore, addressing the plasticity-stability paradox becomes a fundamental requirement for life-long learning machines: achieving a delicate equilibrium between acquiring new knowledge and preserving stability.

Albeit recent work might be promising to mitigate this problem in supervised learning settings \cite{depasquale2018full,ingrosso2019training,capone2019sleep,wilmes2023dendrites,capone2023beyond}, this remains an open issue in reinforcement learning.

Policy-gradient methods utilize estimators of the policy gradient, commonly computed through stochastic gradient ascent algorithms \cite{pmlr-v48-mniha16}. The gradient estimator typically involves the policy's log probability multiplied by an advantage function estimator. While attempting to perform multiple optimization steps using the same trajectory, in order to optimize sample efficiency, empirical evidence suggests this approach often results in excessively large policy updates, leading to instability.
In Trust Region Policy Optimization (TRPO) \cite{schulman2015trust}, a surrogate objective is maximized while ensuring a bounded policy update size.

Proximal Policy Optimization (PPO) employs probability ratios to maximize the surrogate objective. To mitigate excessively large policy updates, PPO adjusts the objective by penalizing deviations of the probability ratio from 1, thanks to a modified objective function, that is clipped within an interval defined by the probability ratio (see details in the methods section). This approach ensures that the optimization remains conservative, bounding the magnitude of policy updates and stabilizing the learning process. PPO has the same spirit of TRPO, but a simpler implementation.
 
Comparisons between PPO algorithm and existing literature gradient-based algorithms reveal superior performance. Notably, on Atari tasks, PPO demonstrates significantly improved sample efficiency over A2C \cite{pmlr-v48-mniha16} and comparable performance to ACER \cite{wang2016sample}, despite its simpler implementation.

For this reason, in response to the challenge described above, we propose a biologically plausible approximation of proximal policy optimization (PPO) \cite{schulman2017proximal} as a promising solution, and we refer to it as lf-cs. 

The computation of gradients in RL usually requires performing a summation on future states (e.g. to evaluate the total discounted reward). This sum can be retroactively transformed into an eligibility trace, that is a summation on past states, allowing the learning rule to be local in time. Nevertheless, due to the constraints imposed by the clip on the surrogate function, this transformation is not feasible. We overcome this problem by proposing a modification of the clipped surrogate function (see Methods section).

lf-cs offers a novel approach to mitigating the impact of limited data and noise in recurrent spiking networks, providing a robust framework for RL in real-life applications.
Our approach introduces two major advancements.  First, it facilitates the seamless assimilation of new information in a new potential policy without necessitating changes to the current policy, addressing the inherent tension between plasticity and stability in lifelong learning machines. Second, it enables the effective replay of experiences without succumbing to policy divergence, ensuring a more stable and reliable learning process.

In contrast to similar techniques present in the literature \cite{wang2016sample, schulman2015trust}, our proposed methodology boasts the additional advantage of computational efficiency in an online setting, making it well-suited for real-time applications. In this paper we benchmark our approach on the task of learning an intelligent behavior from rewards \cite{mnih2016asynchronous}: winning Atari video games provided by the OpenAI Gym \cite{1606.01540} (The MIT License).
We demonstrate that lf-cs outperforms e-prop \cite{bellec2020} on this task, which is deemed to be the state of the art in biologically plausible reinforcement learning.

\section*{Results}

At the heart of our exploration lies the dilemma of choosing between two distinct approaches to model updates: instantaneous updating of weights at each time step, and the accumulation of statistical data over an extended period for delayed updates. 
One approach is to update the model quickly as new data comes in. This allows the model to adapt rapidly but has a downside: it can be too sensitive to 'noisy' data (data that doesn't represent the overall trend). This sensitivity can lead to overfitting, where the model performs well on its training data but poorly on new, unseen data. In policy gradients, this issue is more pronounced because every change in the model (or 'weight update') directly changes the decision-making rules (or 'policy'), leading to instability due to constant changes in how the model makes decisions.
The other approach is to wait until a lot of data is collected before updating the model. This method is slower but more stable. It reduces the risk of being misled by noisy data and makes the model's learning process more reliable.

An optimal strategy might stand in between: quickly update the future policy, while keeping stable the policy used to acquire the data. However, this would bring errors in the evaluation of the policy gradient, that is an on-policy method, and has to be evaluated on the current policy.
This becomes a major problem when using the memory of the same experience more the once to update the policy (experience replay). In this case, after the first policy update, the policy is different from the one used to sample data.

\subsection*{Surrogate function to learn fast and change slow}

\begin{figure}[t!]
    \centering
    \includegraphics[width=\textwidth]{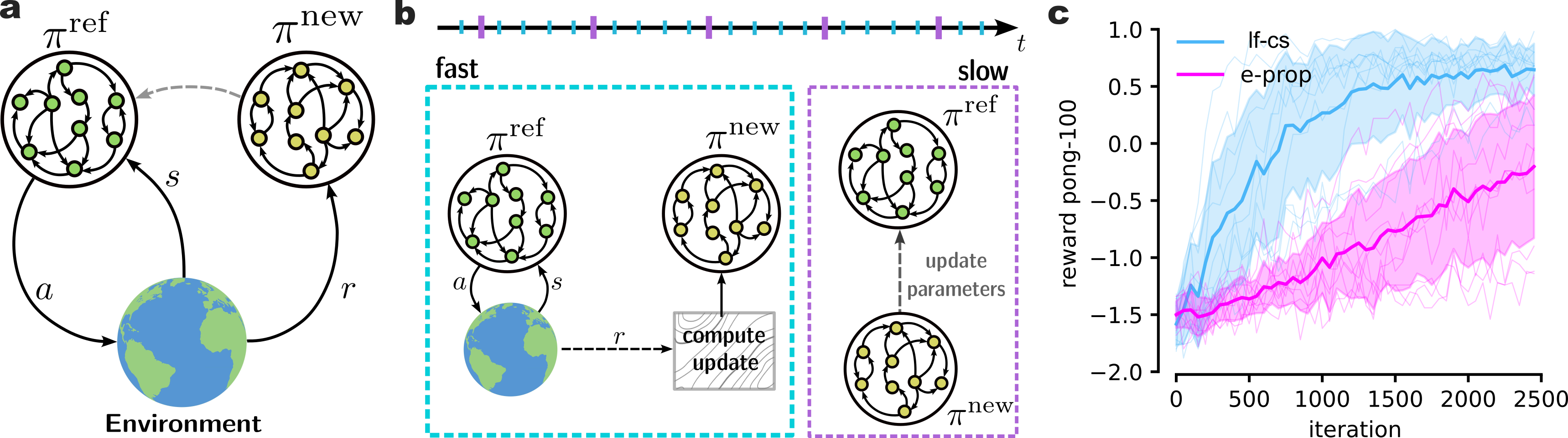}
    \caption{\textbf{Learning on separate timescales.} (\textbf{a}) The model is composed of two networks: a reference policy network $\pi^\mathrm{ref}$ (left) that interacts with the environment receiving states $s$ and emitting actions $a$, and a future policy network $\pi^\mathrm{new}$ (right) that quickly accumulates knowledge based on the new acquired experience and the reward signal $r$. (\textbf{b}) The two networks work on different timescales. During the fast phase (left box, cyan) the reference policy network $\pi^\mathrm{ref}$ interacts with the environment, producing a series of reward signals. This policy is held fixed. The reward signals is used by the second network $\pi^\mathrm{new}$ to updates its policy. On a slower timescales (right box, magenta) the updated policy $\pi^\mathrm{new}$ is transferred into the reference policy $\pi^\mathrm{ref}$. (\textbf{c}) Average reward on the \textsf{Pong-100} environment as a function of the number of interactions with the environment. Separating the timescales in a fast and slow component (cyan) has a beneficial effect on learning speed as compared to only using a single slow timescale (magenta), which is equivalent to the e-prop learning algorithm \cite{bellec2020}. Solid lines are averages over $10$ independent experiments. Shaded areas span the $\pm \mathrm{std}$ intervals. Thin lines of respective colors are individual runs.
    }
    \label{FIGURE:Synaptic_Plasticity}
\end{figure}

We approach this problem by proposing learning fast changing slow (lf-cs), a biologically plausible learning algorithm, inspired to proximal policy optimization (PPO) \cite{schulman2017proximal}. To address the conflicting requirements of data efficiency, i.e. leveraging limited amounts of environment interactions, and stability, we decompose learning into two separate timescales $\tau_f$ and $\tau_s$, which can be thought to be implemented at the network level by two separate modules, which we refer to as $\pi^\mathrm{ref}$ and $\pi^\mathrm{new}$ (see Figure \ref{FIGURE:Synaptic_Plasticity}a). The reference network $\pi^\mathrm{ref}$ interacts with the external environment and at each point in time implements the current agent policy. It emits agent actions $a^t$ and receives updated environment states $s^t$. The environment reward signal $r^t$, which is critical for learning, is fed to the second network $\pi^\mathrm{new}$, which computes the weight update online and embodies the fast timescale $\tau_f$ (see Figure \ref{FIGURE:Synaptic_Plasticity}b left, cyan box). The reference network $\pi^\mathrm{ref}$ is frozen during its interactions with the external environment, ensuring policy stability, and it's updated on a slower timescale $\tau_s$ when the $\pi^\mathrm{new}$ parameters are transferred to the reference network $\pi^\mathrm{ref}$ (see right magenta box in Figure \ref{FIGURE:Synaptic_Plasticity}b). This conceptual framework interpolates between the two regimes of fast online updates ($\tau_f = \tau_s$) and batch learning ($\tau_f \ll \tau_s$), with the optimal configuration often balancing between these two extremes.

We define a surrogate loss function, so that the policy gradient can be evaluated on the data sampled by the reference network (see Methods for details):

\begin{equation}
\mathcal{L} = \mathbb{E}_t \left[ \rho^t R^t \right], \qquad \rho^t(\theta) = \frac{\pi^t_{\theta}(a^t|s^t) }{\pi_{\theta_{\mathrm{old}}}^t(a^t|s^t)}, \qquad R^t = \sum_{{\tau}\geq t} r^{{\tau}} \gamma^{{\tau}-t}
\end{equation}

\noindent where $R^t$ is the total future reward (or return), $r^\tau$ is the reward emitted by the environment at time $\tau$ and $\gamma = 0.98$ is the discount factor.

However, the surrogate loss might bring to numeric errors when the new policy is to large, for this reason it is necessary to introduce a policy regularization so that the new policy is not too far from the reference one. To this purpose PPO introduces the following surrogate loss function

\begin{equation}
\mathcal{L}^{\mathrm{clip}} = \mathbb{E}_t \left[ \mathrm{min} \left( \rho^t R^t,  \mathrm{clip} \left( \rho^t , 1 - \epsilon, 1 + \epsilon \right) R^t \right) \right]
\end{equation}

where $\epsilon$, define the tolerance of the distance between the new and the old policy. However, the non-linearity of the clipping does not allow for evaluating the weight updates online. 
For this reason, we define $\rho_\mathrm{clip}^t = \mathrm{clip} \left(\rho^t,1-\epsilon,1+\epsilon \right)$ and further simplify such expression as follows:

\begin{equation}
\mathcal{L}^{s - \mathrm{clip}} = \mathbb{E}_t \left[ \mathrm{clip} \left( \rho^t , 1 - \epsilon, 1 + \epsilon \right) R^t \right] = \mathbb{E}_t \left[ \rho^t_{\mathrm{clip}} R^t \right]  
\end{equation}

This allows to write explicitly a local (in space and time) synaptic plasticity rule (see Methods for details). We considered a classical benchmark task for learning intelligent behavior from rewards \cite{mnih2016asynchronous}: winning Atari video games provided by the OpenAI Gym \cite{1606.01540} (The MIT License).
We considered the case of a limited temporal horizon, $T=100$. In this case, in the best scenario the agent scores $1$ point and in the worst one the opponent scores $2$ points. The world variables $\xi_k^t \in \mathbb{R}^4$ represent the paddles and ball coordinates.
To win such a game, the agent needs to infer the value of specific actions even if rewards are obtained in a distant future. In fact, learning to win Atari games is a serious challenge for RL even in machine learning. Standard solutions require experience replay (with a perfect memory of many frames and action sequences that occurred much earlier) or an asynchronous training of numerous parallel agents sharing synaptic weight updates \cite{mnih2016asynchronous}.

We demonstrate that, in this environment, the separation of timescales into a fast $\tau_f$ and slow $\tau_s$ component is beneficial to learning speed (see comparison of cyan and magenta curves in Figure \ref{FIGURE:Synaptic_Plasticity}c) as it combines the benefits of efficient data use thanks to the $\pi^\mathrm{new}$ network updating its internal parameters at every interaction with the external environment (each frame), enabling faster learning, with the improved learning stability ensured by the slower reference network $\pi^\mathrm{ref}$, that is updated when the game ends (every $100$ frame). In particular our proposed algorithm (lf-cs) (cyan curve in Figure \ref{FIGURE:Synaptic_Plasticity}c) updates the new network $\pi^\mathrm{new}$ online, i.e. at every interaction with the environment, transferring the synaptic updates to the reference network $\pi^\mathrm{ref}$ at the end of the episode. This configuration leads to significant learning speedups when compared with the standard \emph{e-prop} learning algorithm \cite{bellec2020} (magenta curve in Figure \ref{FIGURE:Synaptic_Plasticity}) where there is no timescales separation and the $\pi^\mathrm{new}$ network accumulates gradients that are then applied and transferred to the $\pi^\mathrm{ref}$ network at the end of the episode.

\subsection*{PPO to allow lf-cs on experience replay}

\begin{figure}[tb!]
    \centering
    \includegraphics[width=\textwidth]{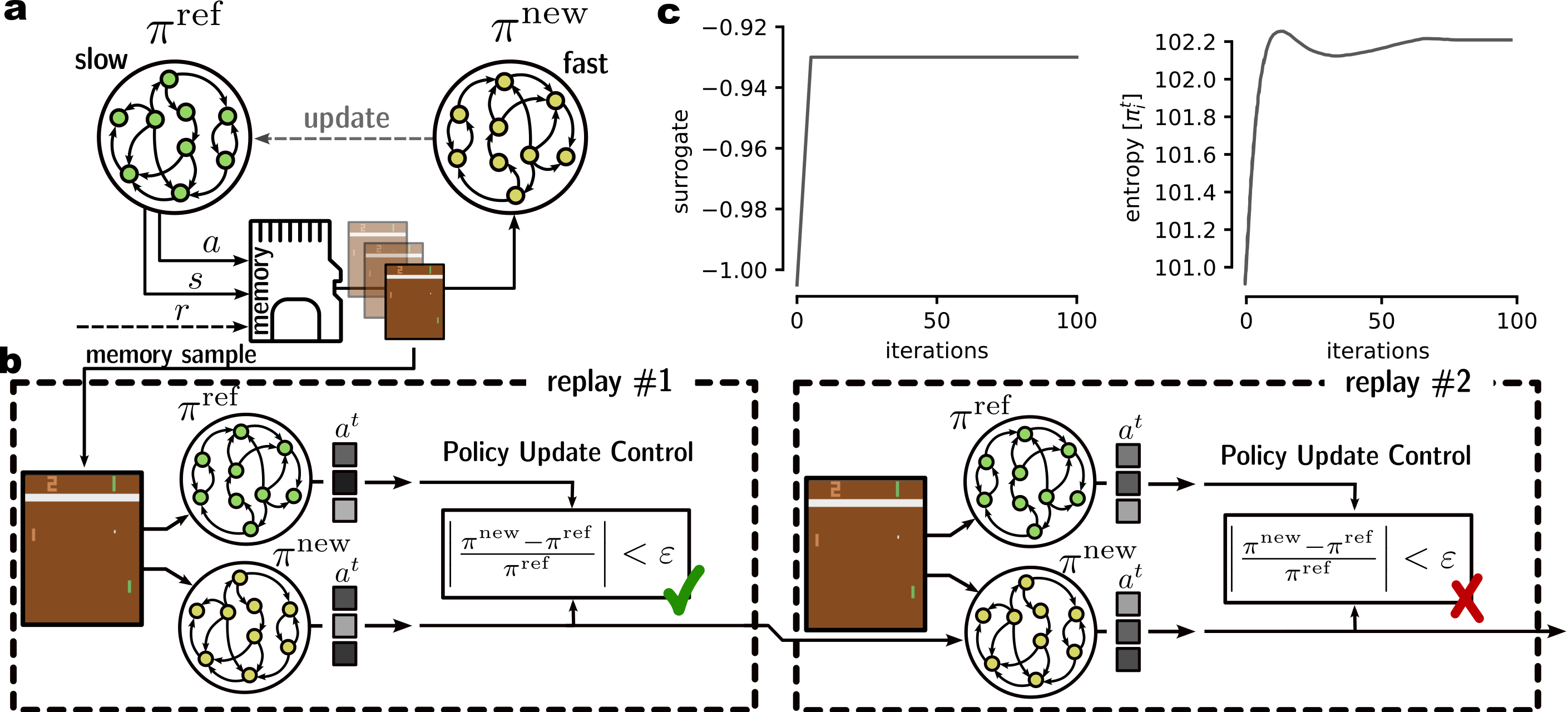}
    \caption{\textbf{Learning via experience replay.} (\textbf{a}) Graphical depiction of the interplay between replay storage-retrieval for the reference policy network $\pi^\mathrm{ref}$ and the fast updating network $\pi^\mathrm{new}$. An external buffers collects the state, action, reward tuple $\left(r^t, s^t, a^t \right)$ generated via the interaction of the reference network $\pi^\mathrm{ref}$ with the environment. The $\pi^\mathrm{new}$ network retrieves the experience from the buffer and updates its internal parameters. The updates are then transferred to the reference network $\pi^\mathrm{ref}$ on a slower timescale. (\textbf{b}) Visual representation of the policy update control mechanism. Given a stiffness $\epsilon$, the policy network $\pi^\mathrm{new}$ computes eligible updates, the magnitude of which (computed as the ratio $\pi^\mathrm{new} / \pi^\mathrm{ref}$) is compared against the update stiffness $\epsilon$. If update is within stiffness constraints, updates are integrated into the reference network $\pi^\mathrm{ref}$, otherwise they are not.  (\textbf{c}) Examples of the policy and entropy evolution over many replays of the same memory. The policy control avoid runaway updates and instabilities due to experience overuse, improving training stability.}
    \label{FIGURE:PPO_Experience_Replay}
\end{figure}

Intuitively, the recollection of past memories boosts agent learning, which can rely on an ever-increasing stock of valuable feedback. 
However, when sampling from the memory bank, experience replay requires a correction to properly evaluate gradients, as described above. 
Our framework allows to replay many time the same episode, without taking the risk of impairing the current policy.
Conceptually, the  reference network $\pi^\mathrm{ref}$ interacts with the environment and assembles the experience memory $\left( a^t, s^t, r^t \right)_t$ which is stored in a memory buffer (see Figure \ref{FIGURE:PPO_Experience_Replay}a). A second network $\pi^\mathrm{new}$ reads from the memory buffer a given experience and computes its internal updates. The two networks $\pi^\mathrm{ref}$ and $\pi^\mathrm{new}$ are then compared and the update magnitude is measured against a pre-defined stiffness parameter $\epsilon$. In particular, the truth value of the following expression $\left| 1 - \frac{\pi^\mathrm{new}}{\pi^\mathrm{ref}} \right| < \epsilon$ determines whether the proposed updates, embodied by the fast-learning network $\pi^\mathrm{new}$, are accepted or not (we refer to Figure \ref{FIGURE:PPO_Experience_Replay}b for a visual representation of the policy update control mechanism). If the updated network $\pi^\mathrm{new}$ differs from the reference network $\pi^\mathrm{ref}$ only slightly, i.e. the policy update condition evaluates to true, the update is accepted and the new network parameters are transferred to the reference network. Vice versa, if the two networks differ substantially after the experience replay, the candidate update is discarded. This control mechanism thus guards against excessive network update magnitudes, restricting the policy to only small, safer improvements and avoiding runaway updates and instabilities due to experience overuse (see Figure \ref{FIGURE:PPO_Experience_Replay}c), leading to reduced variance and overall improvements in learning dynamics.

\subsection*{Role of the stiffness parameter $\varepsilon$}

\begin{figure}[t!]
    \centering
    \includegraphics[width=\textwidth]{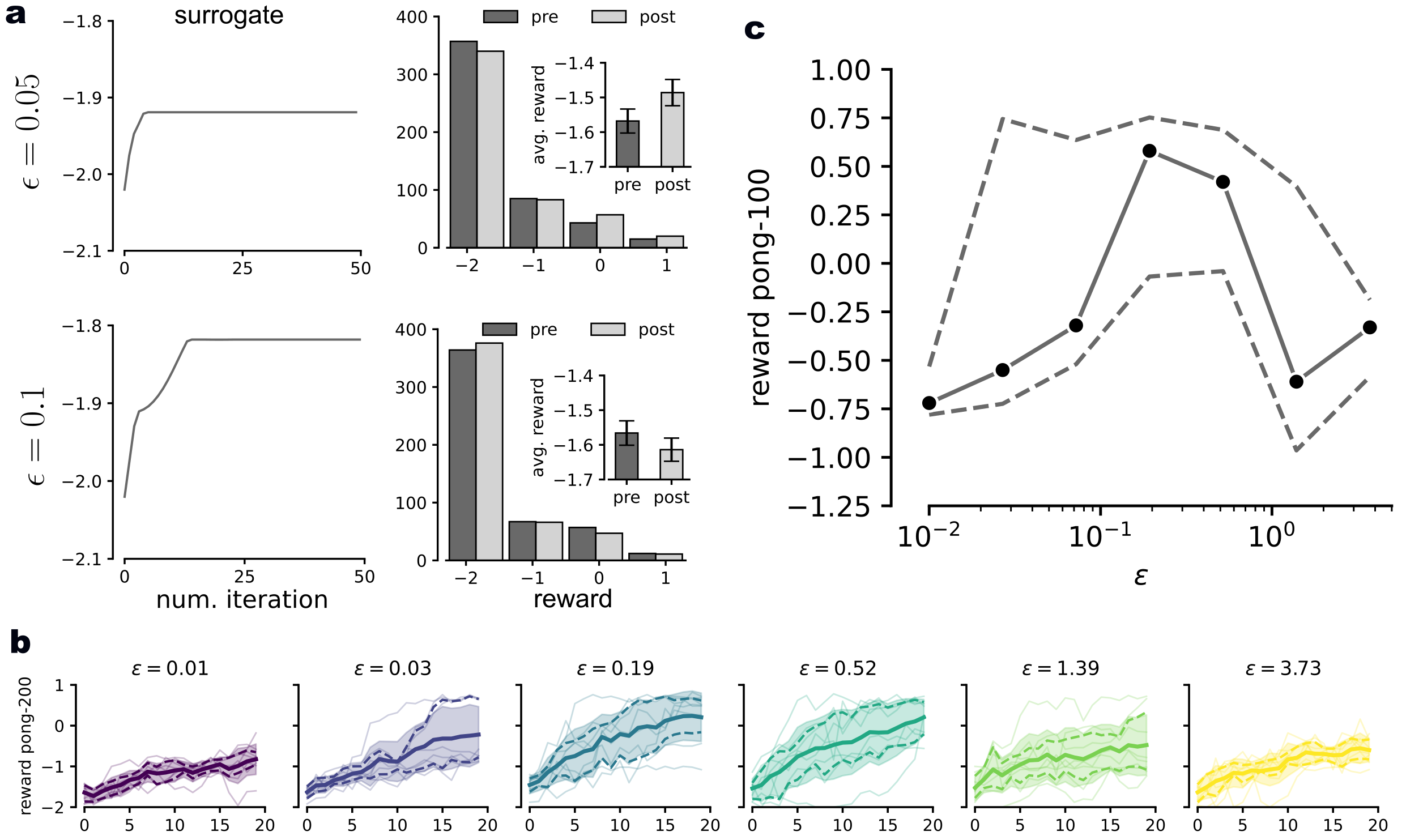}
    \caption{\textbf{Dependence of learning dynamics on stiffness parameter $\varepsilon$.}
    (\textbf{a}). Dynamics of two key metrics: the surrogate function and the final reward $r^t$ pre- and post- policy update, for two choices of the stiffness parameter $\epsilon \in \left\{ 0.05, 0.1 \right\}$. When the policy control update is too large ($\epsilon = 0.1$, right column) major synaptic changes are allowed, resulting in large surrogate function variations (bottom left panel), this degrades overall performances as can be seen by confronting the pre- and post- measured reward (bottom right panel). Better values for the conservation parameter ($\epsilon = 0.05$) ensure sufficient update control (top tow) which benefits performances (top right panel). Inset in the right column panels depict average pre- and post- reward.
    (\textbf{b}). Comparison between performances on the \textsf{atari-pong} environment for different choices of the parameter $\varepsilon$, ranging from no control (right, yellow line, high $\varepsilon$) to strict control (left, purple line, low $\varepsilon$). From the reward profiles a clear optimal $\varepsilon$ regime emerges for intermediate values of the stiffness parameter ($\varepsilon \simeq 0.2$). Thick solid lines report averages over $8$ repetitions, dashed lines mark the $20$- and $80$-percentile, shaded areas cover the $\pm$STD regions, while thin lines are individual runs.
    (\textbf{c}) Maximum reward achieved by an agent in the \textsf{atari-pong} (environment as a function of the $\varepsilon$ parameter. Optimal performances are achieved for intermediate choices of the control parameter. Solid line represent average performance over $10$ independent experiment repetition, while dashed lines represent the $20$-th and $80$-th percentile.}
    \label{FIGURE:Dependence_On_Epsilon}
\end{figure}

A critical hyper-parameter in the lf-cs policy update control mechanisms is the maximum update stiffness $\epsilon$. The magnitude of this hard cutoff determines the maximum allowed network parameters update and should be carefully balanced based on the following trade-off. When the policy control update is too strict, the surrogate network $\pi^\mathrm{new}$ is constrained to be identical to the reference one, as all proposed updates are discarded and learning halts. 
On the other end of the spectrum, i.e. no policy update control is enforced, effectively all suggested updates from experience replays are implemented. This quickly destabilize learning, leading to severe over-fitting on past memories. Moreover, the unbounded optimization of the surrogate function brings to numerical errors, impairing the value of the actual loss function. An example of this is reported in Figure \ref{FIGURE:Dependence_On_Epsilon}a, in which we consider many replays of the same memory, for two different values of $\epsilon$ (top and bottom). The surrogate function is less constrained for larger $\epsilon$ values (top row), resulting in a larger change of the policy, impairing the total reward, as reported in the last row of Figure \ref{FIGURE:Dependence_On_Epsilon}a. For a better choice of the stiffness parameter $\epsilon$ (top row in Figure \ref{FIGURE:Dependence_On_Epsilon}a) the increase of the surrogate function is smaller. This allows for a controlled increase of the total reward after many replays of the memory (right column, top panel, in Figure \ref{FIGURE:Dependence_On_Epsilon}a). The beneficial effects of a balanced stiffness choice are highlighted in Figure \ref{FIGURE:Dependence_On_Epsilon}b where example reward trajectories for an agent interacting with the \textsf{atari-pong} environment as reported for several possible values of $\epsilon$, with intermediate values achieving the best performances. Indeed, when the maximum achieved reward is visualized as a function of the stiffness parameter $\epsilon$ (see Figure \ref{FIGURE:Dependence_On_Epsilon}c), optimal values are revealed to lie in an intermediate region, not too small, not too large.

\begin{figure}
    \centering
    \includegraphics[width=\textwidth]{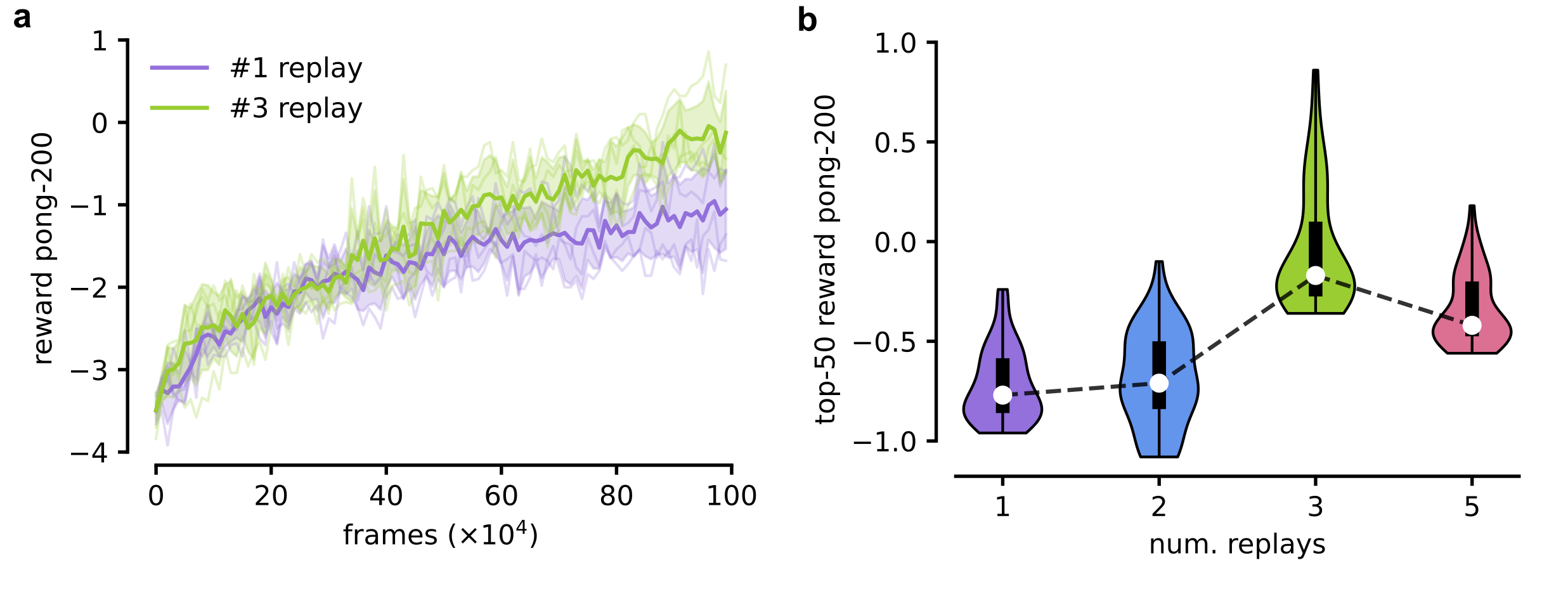}
    \caption{\textbf{Generalization to other conditions.}
    (\textbf{a}) Measured reward in the \textsf{pong-200} Atari environment (game with a temporal horizon of $200$ frames) as a function of the frame numbers for two learning configurations employing a different number of experience replays. Solid thick lines are average over $4$ repetitions, while shaded areas are $\pm$STDs. Thin lines are individual runs.
    (\textbf{b}) Violin plot reporting the distribution of the top-50 measured rewards across experiments for different numbers of experience replays. Thick whiskers represent the $25$- and $75$-percentile, solid white markers report the median.
    \label{FIGURE:Generalization_to_Other_Conditions}
    }
\end{figure}

\subsection*{Generalization to longer temporal horizons}

As a first attempt to demonstrate the capability of the method to other conditions, we consider a slightly harder situation in which the temporal duration of the \textsf{atari-pong} game is $200$ frames. In this case the maximum score is $2$ and the minimum is $-5$. We let our agent interact with the environment and tested our proposed learning algorithm with a variable number of memory replays, from $1$ up to $5$ replays per memory. In Figure \ref{FIGURE:Generalization_to_Other_Conditions}a we reported the reward traces measured for our agent, color-coded for increasing number of memory replays. We observed a substantial increase in learning speed as a function of additional replays, signaling the beneficial contribution of the lf-cs policy-control mechanism on learning dynamics. Indeed, the distributions of the top-$50$ measured episode rewards as a function of the number of memory replays \ref{FIGURE:Generalization_to_Other_Conditions}b reveal a substantial distribution shift towards higher rewards for an increasing number of replays,
providing compelling evidence for the virtuous interplay of learning from memory replays supported by the lf-cs policy update control mechanism. 

\section*{Methods}

\subsection*{Recurrent Spiking Network}

We consider a Recurrent Spiking Network (RSA) defined as follows. Each network $\alpha$ is composed of $N=500$ neurons. Neurons are described as real-valued variable $v_{i}^t \in \mathbb{R}$, where the $i \in \left\{1, \dots, N\right\}$ label identifies the neuron and $t \in \left\{1, \dots, T\right\}$ is a discrete time variable. Each neuron exposes an observable state $s_{i}^t \in \left\{0, 1 \right\}$, which represents the occurrence of a spike from neuron $i$ of the module $\alpha$ at time $t$. 
We then define the following dynamics for our model: 

\begin{equation}
\left \{
\begin{array}{l}
\hat s_{i}^{t} = \left( 1-\frac{\Delta t}{\tau_s} \right) \, \hat s_{i}^{t-1}  + \frac{\Delta t}{\tau_s} \, s_{i}^{t} \\

v_{i}^{t} = \left( 1-\frac{\Delta t}{\tau_m} \right) \, v_{i}^{t-1} + \frac{\Delta t}{\tau_m} \, \left( \sum_j  w_{ij}^{\alpha} \hat s_{j}^{t-1} + I_{i}^t + v_\mathrm{rest} \right) -w_\mathrm{res} s_{i}^{t-1}\\

s_{i}^{t+1} = \Theta \left[ v_{i}^t - v^\mathrm{th}  \right]
\end{array}
\right.
\end{equation}

\noindent where $\Delta t$ is the discrete time-integration step, while $\tau_s$ and $\tau_m$ are respectively the spike-filtering time constant and the temporal membrane constant, $\Theta$ is the Heaviside step function. Each neuron is a leaky integrator with a recurrent filtered input obtained via a synaptic matrix $ w_{ij}$ (from neuron $j$ to neuron $i$) and an external input current $I^t_{i}$. The $w_\mathrm{res} = 20$ parameter accounts for the reset of the membrane potential after the emission of a spike, while $v^\mathrm{th}=0$ and $v_\mathrm{rest}=-4$ are the threshold and the rest membrane potential. The input to the network is a random projection of the state $\xi_h^t \in \mathbb{R}^d$ describing the environment: $I_{i}^t = \sum_h w_{ih}^{\mathrm{in}} \xi_h^t $. 

The environment variables $\xi_k^t$ represent the paddles and ball coordinates (the dimensionality of the world is $d=4$, see also \cite{capone2022towards}).

The elements in $w_{ih}^{\mathrm{in}}$ are randomly extracted from a zero mean Gaussian distribution and variance $\sigma^{\mathrm{in}}$. We refer to Table \ref{tab:model_parameters} for a full list of model parameters used in the experiments. The action is one-hot encoded: $\mathds{1}_{k}^t$ at time $t$ assumes the value $1$ if and only if $a^t = k$, and $0$ otherwise. Finally, we introduce the following quantities, that are relevant to the learning rules described in the following sections: the pseudo-derivative $p_i^t$ and the spike response function $e_j^t$. Their respective definitions are:

\begin{equation}
p_i^t \coloneqq \frac{e^{v_i^t/\delta v}}{\delta v (e^{v_i^t/\delta v}+1)^2}, \qquad e_{j} ^t \coloneqq \frac{\partial v_{i}^t}{\partial w_{ij}^{\alpha}}.
\label{EQ:pseudo-derivative-and-spike-response-definition}
\end{equation}

\noindent The pseudo-derivative $p_{i}^t$ is defined similarly to \cite{bellec2020,capone2022error} and peaks at $v_i^t = 0$, while $\delta v$ defines its width. The spike response function $e_{j}^t$ can be computed iteratively as (fixing $e_j^{t=0} = 0$):

\begin{equation}
e_{j} ^{t+1}   = \exp{ \left(- \frac{\Delta t}{\tau_m}\right)} \, e_{j} ^t  +\left( 1 - \exp{ \left(-\frac{\Delta t}{\tau_m} \right) }  \right) \, \hat{s}^{t}_{i}.
\label{dv_dynamics}
\end{equation}

\begin{table}[]
    \centering
    \begin{tabular}{cccccccccc}
        $N$ & $\Delta t (ms)$ & $\tau_s$ & $\tau_m$ & $v_\mathrm{rest}$ & $w_\mathrm{res} (mV)$ & $v^\mathrm{th} (mV)$ & $\sigma^\mathrm{in}$ & $\delta v (mV)$ & $\gamma$ \\ \hline\hline \\[-7pt]

        $500$ & $1 \times 10^{-3}$ & $4$ $\Delta t$ & $6$ $\Delta t$ & $-4$ & $20$ & $0$ & $10$ & $0.05$ & $0.98$ \\[3pt] \hline
    \end{tabular}
    \caption{Model parameters}
    \label{tab:model_parameters}
\end{table}

\subsection*{Actor-Critic in RSNs}

The policy $\pi_k^t$ defines the probability to pick the action $a^t = k$ at time $t$, and it is defined as:

\begin{equation}
\pi_k^t \coloneqq \frac{\exp{(y_k^t)}}{\sum_h \exp{(y_h^t)}},
\end{equation}

\noindent where $y_k^t = \sum_i A_{ki}^{\pi} \hat{s}_{i}^t$ is a linear readout of the (filtered) network spiking activity $\hat{s}_{i}^t$ via a readout matrix $A_{ki}^\pi$. Following a policy gradient approach to maximize the total reward obtained during the episode, it is possible to write the following loss function (see \cite{bellec2020,sutton2018reinforcement}):

\begin{equation}
\mathcal{L}_A = - \sum_t R^t \mathrm{log} \left( \pi_k^t \right),
\label{pg_loss}
\end{equation}

\noindent where $R^t = \sum_{{\tau}\geq t} r^{{\tau}} \gamma^{{\tau}-t}$ is the total future reward (or return), $r^t$ is the reward emitted by the environment at time $t$ and $\gamma = 0.99$ is the discount factor. In practice, the gradient of $\mathcal{L}_A$ is known to have high variance and the efficiency of the learning algorithm can be improved using the actor-critic variant of the policy
gradient algorithm. This requires an additional network output $V^t$, defined as $V^t = \sum_k C^V_k \hat{s}_k^t$, with $C_k^V$ the dedicated readout matrix, which should predict the value function $\mathcal{L}_A$ and represents the critic in the actor-critic framework. We can define an additional loss function:

\begin{equation}
\mathcal{L}_C = \frac{1}{2} \sum_{t} \left( R^t - V^t \right),
\label{EQ:Critic_PG_Loss}
\end{equation}

\noindent that measures the accuracy of the value estimate. As $V^t$ is independent of the action $a^t$ when conditioned on the network activity, one can show that the expected change induced by the presence of the critic vanishes, i.e.:

\begin{equation}
\mathbb{E} \left[ V^t \frac{d}{d w_{ij}} \log{\left( \pi \left( a^t | y^t \right) \right)} \right] = 0.
\label{EQ:Critic_Expected_Value_Zero}
\end{equation}

One can thus use the critic network to define a new estimator of the loss gradient with reduced variance, by replacing the expected future reward $R^t$ with the advantage $A^t = R^t-V^t$ in Eq. \eqref{pg_loss}. We obtain the following loss:

\begin{equation}
\mathcal{L}_{AC} = - \sum_t \left( R^t - V^t \right) \mathrm{log} \left( \pi_k^t \right) + \lambda_C \mathcal{L}_C,
\label{ac_loss}
\end{equation}

\noindent where $\lambda_C$ is a scalar that balances the trade-off between actor proficiency and critic prediction precision. We note here that, despite our formulation is general for the actor-critic framework, in the presented experiments we set $V^t$ and $\lambda_C$ to $0$. The loss formulation in \eqref{ac_loss} is not physically realistic as network updates in the present require future knowledge via the return $R^t$. To surpass this difficulty, standard derivations introduce the temporal difference error $\delta^t = r^t + \gamma V^{t+1} - V^t$, which allows the following re-writing:

\begin{equation}
R^t - V^t = \sum_{{\tau}\geq t} \delta^{{\tau}} \gamma ^{{\tau}-t}.
\label{EQ_Temporal_Difference_Error}
\end{equation}

Using the one-hot encoded action $\mathds{1}^t_k$ at time $t$, which assumes the value $1$ if and only if $a^t = k$ (else it has value $0$), we arrive at the following synaptic plasticity rules (the derivation is the same as in \cite{bellec2020}):

\begin{equation}
\left \{
\begin{array}{l}
\Delta w_{ij} \propto \dfrac{d \mathcal{L}_{AC}}{d w_{ij} } \simeq  \sum_k  \delta^t  A_{ik}^{\pi} \sum_{{\tau} \leq t} \gamma^{t-{\tau}} \left(  \mathds{1}_k^\tau - \pi_k^{{\tau}} \right) p_{i}^{{\tau}} e_{j} ^{{\tau}} + \lambda_C \sum_k  \delta^t  C_{k}^{V} \sum_{{\tau} \leq t} \gamma^{t-{\tau}} p_{i}^{{\tau}} e_{j} ^{{\tau}}\\[10pt]

\Delta A_{ki}^{\pi} \propto \dfrac{d \mathcal{L}_{AC}}{d A_{ki}^{\pi} } = \delta^t  \sum_{{\tau} \leq t} \gamma^{t-{\tau}} \left( \mathds{1}_k^\tau - \pi_k^{{\tau}} \right) \hat{s}_{i}^{{\tau}}\\[10pt]

\Delta C_{k}^{V} \propto \dfrac{d \mathcal{L}_{AC}}{d C_{ik}^{V} } =  \delta^t  \sum_{{\tau} \leq t} \gamma^{t-{\tau}} \hat{s}_{k}^{{\tau}}
\end{array}
\right.
\label{eq_AC_from_Bellec}
\end{equation}

Intuitively, given a trial with high rewards, the policy gradient changes the network output $y_k^t$ to increase the probability of the actions $a^t$ that occurred during this trial. Even in this case, the plasticity rule is local both in space and time, allowing for an online learning for the agent-network. All the weight updates are implemented using the Adam optimizer \cite{kingma2014adam} with default parameters and learning rate $1 \times 10^{-3}$. In the experiment reported in the paper, the value function is set to zero, this is equivalent to the policy gradient, and results in replacing $\delta^t$ with the instantaneous reward $r^t$.

\subsection*{Policy regularization}

While it is appealing to perform multiple steps of optimization on this loss using the same trajectory, doing so is not well-justified, and empirically it often leads to destructively large policy updates. In TRPO \cite{schulman2015trust}, an objective function, the surrogate objective, is maximized subject to a constraint on the size of the policy updates. PPO proposed a much simpler framework to obtain the same effect: limiting the size of policy update. Let $\rho^t \left( \theta \right)$ denote the ratio:

\begin{equation}
\rho^t \left( \theta \right) = \frac{\pi^t_{\theta} \left( a^t |s^t \right)}{\pi_{\theta_{\mathrm{old}}}^t \left(a^t |s^t \right)}.
\end{equation}

We can define the surrogate loss-function as: $\mathcal{L}^\mathrm{clip} = \mathbb{E}_t \left[ \rho^t \left( \theta \right)  A^t \right]$, where $A^t$ is the advantage. PPO introduces the following regularization (clip):

\begin{equation}
\mathcal{L}^{\mathrm{clip}} = \mathbb{E}_t \left[ \mathrm{min} \left( \rho^t A^t,  \mathrm{clip} \left( \rho^t , 1 - \epsilon, 1 + \epsilon \right) A^t \right) \right].
\end{equation}

However, the non-linearity of the clipping does not allow for evaluating the weight updates online. This makes it indeed impossible to derive an expression that is equivalent to Eq.\ref{eq_AC_from_Bellec}. For this reason, we define $\rho_\mathrm{clip}^t = \mathrm{clip} \left(\rho^t, 1 - \epsilon, 1 + \epsilon \right)$ and further simplify such expression as follows:

\begin{equation}
\mathcal{L}^{s - \mathrm{clip}} = \mathbb{E}_t \left[ \mathrm{clip} \left( \rho^t , 1 - \epsilon, 1 + \epsilon \right) A^t \right] = \mathbb{E}_t \left[ \rho^t_{\mathrm{clip}} A^t \right]. 
\end{equation}

\noindent This allows for deriving the following online plasticity rules (The rules for the value function are left unchanged from last equation in \ref{eq_AC_from_Bellec})):

\begin{equation}
\left \{
\begin{array}{l}
\Delta w_{ij} \propto \dfrac{d \mathcal{L}^{s - \mathrm{clip}}}{d w_{ij}^{A}} \simeq  \sum_k  \delta^t  A_{ik}^{\pi} \sum_{{\tau} \leq t} \gamma^{t-{\tau}} \rho^{{\tau}}_{\mathrm{clip}} \left(  \mathds{1}_k^\tau - \pi_k^{{\tau}} \right) p_{i}^{{\tau}} e_{j} ^{{\tau}}\\

\Delta A_{ki}^{\pi} \propto \dfrac{d \mathcal{L}^{s - \mathrm{clip}}}{d A_{ki}^{\pi} } = \delta^t  \sum_{{\tau} \leq t} \gamma^{t-{\tau}} \rho_{\mathrm{clip}}^{{\tau}} \left( \mathds{1}_k^\tau - \pi_k^{{\tau}} \right) \hat{s}_{i}^{{\tau}}
\end{array}
\right.
\label{eq_AC}
\end{equation}

\section*{Discussion}

The proposed approach involves a local (in space and time) approximation of Proximal Policy Optimization (PPO) directly implementable in recurrent spiking neural networks. This entails the use of two parallel networks: one acquires new data with a stable policy, while the other quickly updates parameters to evaluate the future policy. The approach aims to improve performances and stability in learning within a recurrent spiking neural network \cite{wang2023evolving,sankaran2022event}.

The key advancements of this approach are twofold. First, it allows the seamless assimilation of new information without requiring changes to the current policy, addressing the tension between plasticity and stability in lifelong learning machines. Second, it enables the effective replay of experiences without succumbing to policy divergence, ensuring a more stable and reliable learning process. Unlike other experience replay (ER) techniques, this methodology offers the additional advantage of computational efficiency in an online setting, making it well-suited for real-time applications \cite{chevtchenko2021combining}.
This innovation contributes to the growing body of research aimed at enhancing the practicality and efficiency of Reinforcement Learning (RL) algorithms in diverse and dynamic environments. The paper aims to shed light on the optimal strategies for online learning in policy gradient methods, considering the balance between immediate adaptability and long-term stability in ever-evolving environments. The results not only showcase the effectiveness of the proposed methodology but also highlight its potential impact on real-world applications. This work contributes to the ongoing discourse on the intersection of RL, computational neuroscience, and artificial intelligence, pushing the boundaries of what is achievable in the realm of autonomous, adaptive systems \cite{shen2023brain,capone2022error}.

\section*{Acknowledgments}
This work has been cofunded by the European Next Generation EU grants CUP B51E22000150006 (EBRAINS-Italy IR00011 PNRR Project).

We acknowledge the use of Fenix Infrastructure resources, which are partially funded from the European Union’s Horizon 2020 research and innovation programme through the ICEI project under the grant agreement No. 800858.

\bibliographystyle{unsrt}

\clearpage
\appendix
\renewcommand{\thefigure}{S\arabic{figure}}
\setcounter{figure}{0} 

\begin{center}
    \Huge Supplementary Materials
\end{center}

\section{Exploration of learning rates}

\begin{figure}[htb]
    \centering
    \includegraphics[scale=1.2]{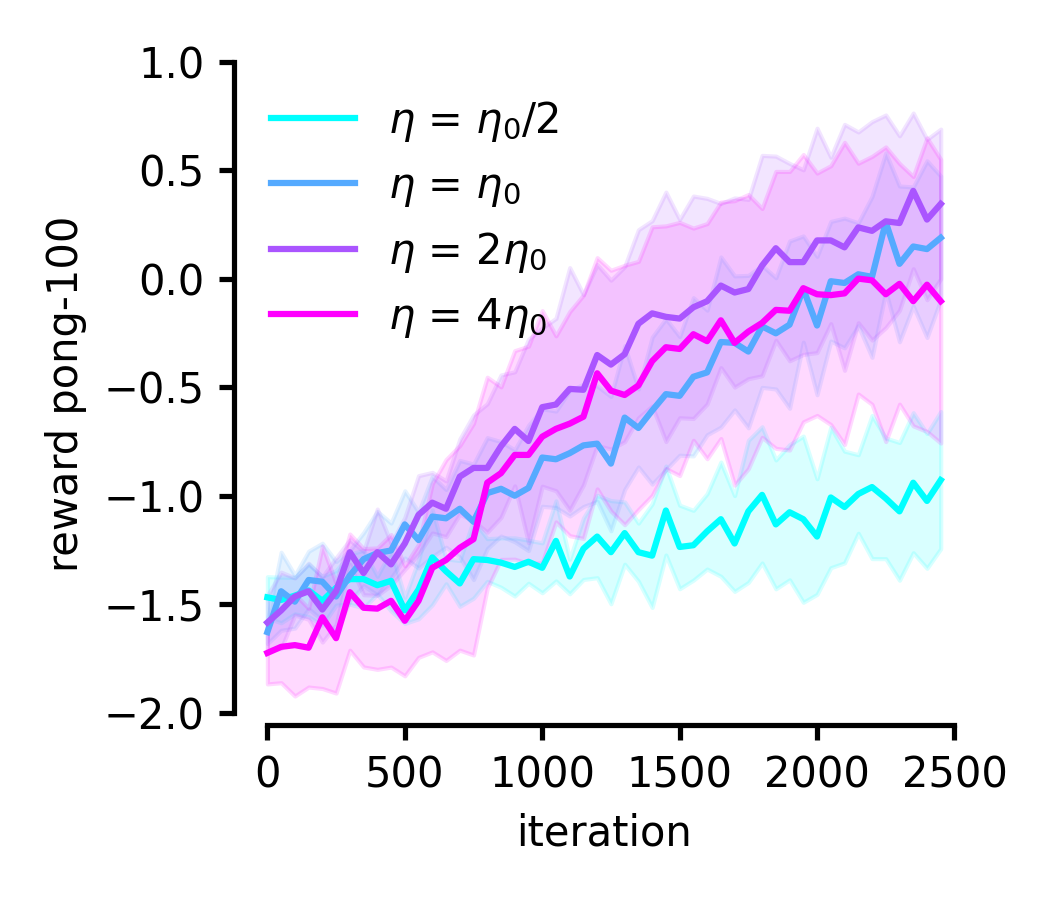}
    \caption{\textbf{Exploration of learning rates} Measured reward in the \textsf{pong-100} environment as a function of the environment interactions. Different colors code for different learning rates $\eta$ used by the Adam optimizer to update the model parameters. We set $\eta_0 = 1.5 \times 10^{-3}$. Solid lines are averages over $5$ repetitions, while shaded areas represent the $\pm \mathrm{std}$.}
    \label{fig:eta-search}
\end{figure}

We explore a range of learning rates to validate the robustness of our results against variation of this hyper-parameter. We fixed $\eta_0 = 1.5 \times 10^{-3}$ and trained four different agents on the \textsf{pong-100} environment exploring the following set of learning rates $\eta \in \left\{ \eta_0 / 2, \eta_0, 2 \eta_0, 4 \eta_0 \right\}$. For each experiment we run $5$ repetitions and reported the measured averages (see Figure \ref{fig:eta-search}). We found that agent performances are stable for a broad range of the explore parameters.

\section{Importance of recurrent connections}

\begin{figure}[htb]
    \centering
    \includegraphics[scale=1.2]{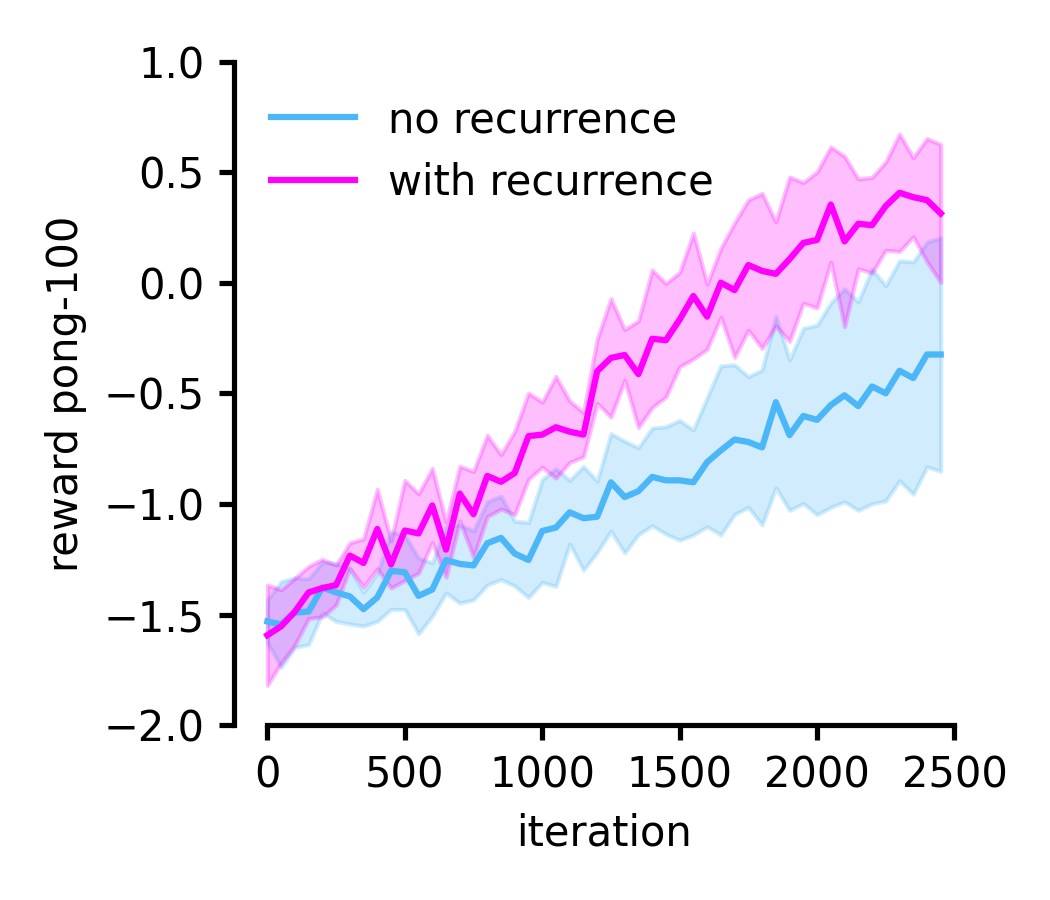}
    \caption{\textbf{Importance of recurrent connections}. Measured reward in the \textsf{pong-100} environment as a function of the environment interactions for two agents. The first agent (cyan) has no recurrent connection ($J_\mathrm{rec} \equiv 0$), while the second agent (magenta) has a non-zero, learnable recurrent synaptic matrix. Solid lines are averages of $10$ repetitions, while shaded areas represent the $\pm \mathrm{std}$ intervals.}
    \label{fig:recurrence}
\end{figure}

To validate the importance of recurrent connections we perform a validation experiment in which the recurrent synapses were severed by manually setting $J_\mathrm{rec} \equiv 0$ (and turning off any learning for this parameter) and only trained the readout connections $J_\mathrm{out}$, and compared this setting with the standard configuration in which both recurrent and readout connections are non-zero and trained. The results of this comparison are reported in Figure \ref{fig:recurrence}, where we have shown how performance in the \textsf{pong-100} improve significantly faster when the model can adapt its recurrent connectivity (compare the magenta line with the cyan line in Figure \ref{fig:recurrence}).

\end{document}